\documentclass[10pt,twocolumn,letterpaper]{article}

\usepackage{wacv}
\usepackage{times}
\usepackage{epsfig}
\usepackage{graphicx}
\usepackage{amsmath}
\usepackage{amssymb}
\usepackage{booktabs} 
\usepackage{multirow}
\usepackage{enumitem} 
\usepackage{makecell}
\usepackage[bookmarks=false]{hyperref}

\usepackage[symbol]{footmisc}

\wacvfinalcopy 

\def\wacvPaperID{****} 

\ifwacvfinal\fi
\begin{document}

\title{Graph Neural Networks for Image Understanding Based on Multiple Cues: Group Emotion Recognition and Event Recognition as Use Cases}


\author{\parbox{16cm}{\centering
    {\large Xin Guo$^1$, Luisa F. Polan\'{i}a$^2$, Bin Zhu$^1$, Charles Boncelet$^1$, and Kenneth E. Barner$^1$}\\
    {\normalsize
    $^1$ Department of Electrical and Computer Engineering, University of Delaware, Newark, DE, USA\\
    Email: \{guoxin, zhubin, boncelet, barner\}@udel.edu\\
    $^2$ Target Corporation, Sunnyvale, California, USA,\\
    Email: Luisa.PolaniaCabrera@target.com }}
}

\maketitle
\ifwacvfinal\thispagestyle{empty}\fi

\begin{abstract}
   A graph neural network (GNN) for image understanding based on multiple cues is proposed in this paper. Compared to traditional feature and decision fusion approaches that neglect the fact that features can interact and exchange information, the proposed GNN is able to pass information among features extracted from different models. Two image understanding tasks, namely group-level emotion recognition (GER) and event recognition, which are highly semantic and require the interaction of several deep models to synthesize multiple cues, were selected to validate the performance of the proposed method. It is shown through experiments that the proposed method achieves state-of-the-art performance on the selected image understanding tasks. In addition, a new group-level emotion recognition database is introduced and shared in this paper. 
\end{abstract}

\section{Introduction}
Deep learning methods have shined in many computer vision tasks~\cite{DBLP:journals/corr/GilmerSRVD17,8373900,DBLP:journals/corr/abs-1709-01507,7298768,DBLP:journals/corr/abs-1711-06396} ever since Krizhevsky \etal~\cite{Krizhevsky:2012:ICD:2999134.2999257} achieved the top classification accuracy on the Large Scale Visual Recognition Challenge (LSVRC)~\cite{ILSVRC15} in 2010. This overwhelming success is due to the ability of deep learning methods to learn at different levels of abstraction from data. Different tasks need different levels of abstraction. For example, tasks such as image segmentation focus on pixel-level information, whereas tasks such as GER and event recognition require a deeper semantic understanding of image contents and the aggregation of information from facial expressions, posture, people layout and background environments~\cite{Dhal15b}. Single deep models are typically sufficient to achieve excellent performance for object recognition and image segmentation tasks while the aggregation of several deep models that extract features not only from the whole image but also from salient areas is typically needed for image understanding tasks ~\cite{Guo:2017:GER:3136755.3143017, Tan:2017:GER:3136755.3143008, Wei:2017:NDF:3136755.3143014}. 

Understanding the meaning and content of images remains a challenging problem in computer vision. Attempts to extract high-level semantic information for image understanding include the work in~\cite{obj14}, which proposes the Object Bank, an image representation constructed from the response of multiple object detectors. Recently, modular networks have been proposed to perform visual understanding tasks by using several reusable and composable modules that carry on different functions~\cite{Masc18}. In a nutshell, the state-of-the-art in image understanding is based on exploiting the principle of compositionality, meaning that a set of entities and their interactions are used to understand an image. 

The aggregation of information from deep models trained on different entities or cues is typically implemented through decision and feature fusion ~\cite{Guo:2017:GER:3136755.3143017, Guo:2018:GER:3242969.3264990,Tan:2017:GER:3136755.3143008, Sun:2016:LDE:2993148.2997640}. However, such methods neglect the fact that features can interact with each other to exchange information. Recurrent neural networks (RNNs) are widely used to aggregate features~\cite{Li:2016:HLP:2993148.2997636,Sun:2016:LDE:2993148.2997640,Wei:2017:NDF:3136755.3143014}, but mostly from the same model since features of different models usually have different size. Another major drawback of RNN-based approaches is that they only consider sequential information but ignore spatial relations between entities present in the image. 

Motivated by addressing the image understanding problem from learning features of multiple cues jointly, we propose a GNN model, which can be seen as a generalization of RNNs from sequential to graph data~\cite{Qi17}. Features from regions of interest corresponding to multiple cues are extracted from the images and used as the nodes of the GNN. The hidden representation of each node evolves over time by exchanging information with its neighbors. One major advantage of the proposed model is its ability to deal with different number of inputs, which is relevant because the number of entities of interest vary between images, \textit{e.g.} the number of faces. Another advantage is that each input is allowed to have a different size, which is important because different entities may have feature representations of different size. The performance of the proposed approach is validated on GER and event recognition tasks.

The models closer to the proposed model are those of \cite{liu18} and \cite{li17b} because they also use graphs to address image understanding tasks. However, the method in \cite{liu18} focuses only on the problem of object detection. The method in \cite{li17b} exploits connections across semantic levels, while the proposed method exploits connections between multiple cues and  between features  belonging  to  the  same  cue type.   The model  in \cite{li17b} also differs from ours in the aggregation  functions that are employed. Also, it does not use RNNs to update the features.

The major contributions of this work are summarized as follows: (1) A GNN model to address the problem of image understanding based on multiple cues. (2) The topology of the graph is dynamic because the number of entities of interest varies between images. Also, the proposed GNN model is able to deal with different number of inputs, where each input is allowed to have a different size. (3) A dataset is introduced to address the GER problem in realistic scenarios. (4) Extensive experiments are conducted to illustrate the performance of the proposed GNN on GER and event recognition tasks. Code and database are available at {https://github.com/gxstudy/Graph-Neural-Networks-for-Image-Understanding-Based-on-Multiple-Cues}.


\section{Related Work}
\label{related_work}

\subsection{Graph Neural Network}
Graph neural networks were first proposed by Gori \etal~\cite{1555942} and detailed in Scarselli~\cite{4700287} as a trainable recurrent message passing network applicable to sub-graph matching, web page ranking, and some toy problems derived from graph theory. Graph neural networks extend the notion of convolution and other basic deep learning operations to non-Euclidean grids~\cite{Meng_2018_ECCV}. In 2015, Li \etal~\cite{DBLP:journals/corr/LiTBZ15} proposed to modify GNNs to use gated recurrent units (GRUs) and modern optimization techniques. Their work showed successful results in synthetic tasks that help develop learning algorithms for text understanding and reasoning~\cite{DBLP:journals/corr/WestonBCM15}. In~\cite{Kipf16}, Kipf and Welling introduced graph convolutional networks as multi-layer CNNs where the convolutions are defined on a graph structure for the problem of semi-supervised node classification. A message passing algorithm and aggregation procedure for GNNs proposed by Glimer~\cite{DBLP:journals/corr/GilmerSRVD17} achieved state-of-the-art results for molecular prediction.  In 2018, Meng \etal~\cite{Meng_2018_ECCV} proposed a GNN model to learn relative attributes from pairs of images. Meanwhile, a GNN model was proposed by Garcia and Bruna~\cite{garcia2017few} to learn valuable information from limited and scarce training samples for image classification. In~\cite{Qi17}, a 3D GNN for RGBD semantic segmentation, which leverages both the 2D appearance
information and 3D geometric relations,  was proposed.

\subsection{Group-level emotion recognition}
 Group-level emotion recognition has gained popularity in recent years due to the large amount of data available on social networks, which contain images of groups of people participating in social events. In addition, GER has applications in image retrieval~\cite{Dhal10}, shot selection~\cite{Dhal15}, surveillance~\cite{Bull05}, event summarization~\cite{Dhal15}, social relationship recognition~\cite{8756602}, and event detection~\cite{Vand15}, which motivates the design of automatic systems capable of understanding human emotions at the group level. Group emotion recognition is challenging due to face occlusions, illumination variations, head pose variations, varied indoor and outdoor settings, and faces at different distance from the camera which may lead to low-resolution face images.
 
\begin{figure}[t]
\begin{center}
   \includegraphics[width=.9\linewidth]{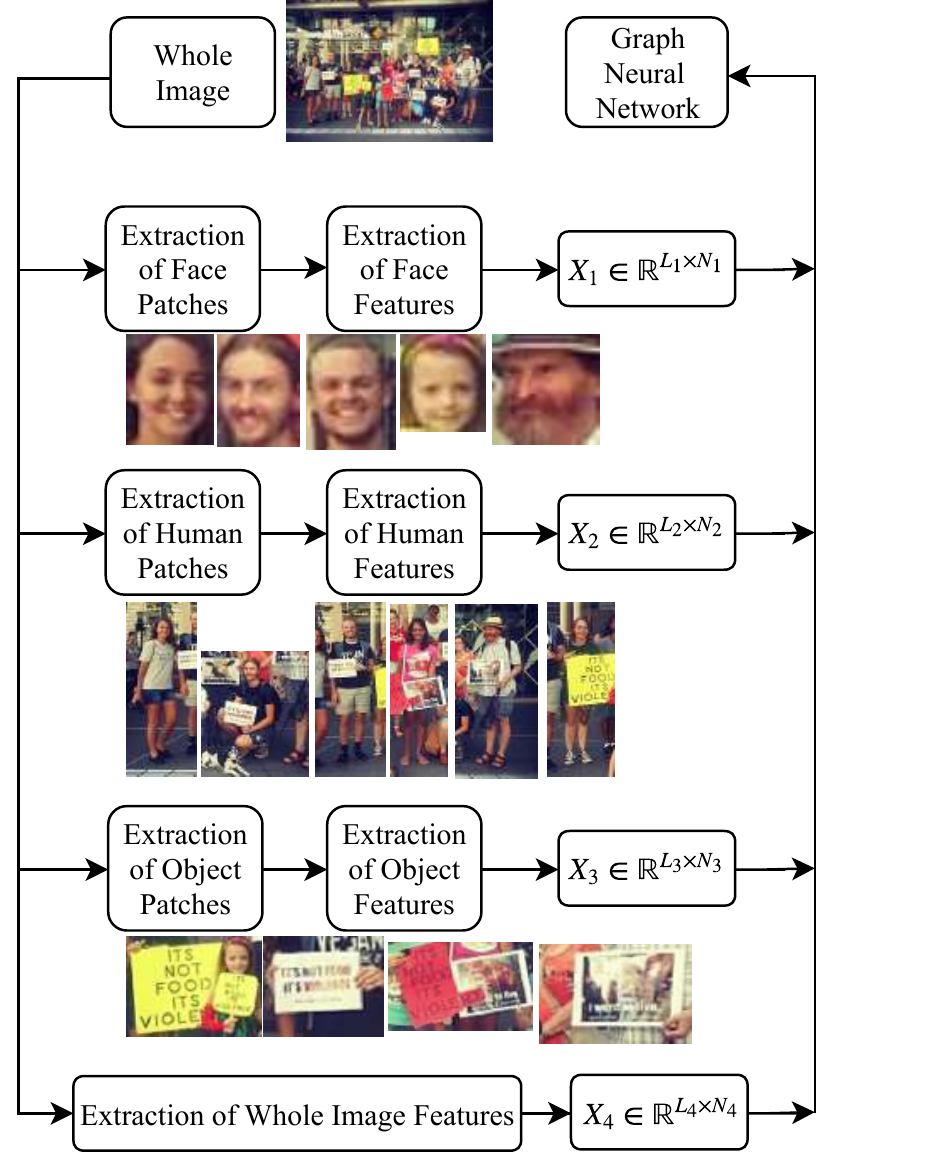}
\end{center}
   \caption{An illustration of how to build a complete graph from an image. Face, human, and object patches are first cropped using different object detection models, then features are extracted using CNN-based models. Each feature vector is a node in the graph.}
\label{fig1}
\end{figure}

 Contextual information is crucial for the GER problem. In Figure~\ref{fig1}, it would be difficult to infer the group emotion by only extracting information from faces, since many of the humans in the image are posing for the photo. However, it is only when contextual information is extracted, in the form of salient objects, such as demonstration posters, that the real emotion of the group is exposed. 
 
 The EmotiW Group-level Emotion Recognition Sub-challenge~\cite{Dhall:2016:ERW:2993148.3007626} was created with the aim of advancing group-level emotion recognition. In this annual sub-challenge, the collective emotional valence state is classified as positive, neutral, or negative using the Group Affect Database 2.0~\cite{Dhall:2016:ERW:2993148.3007626, Dhall:2017:IGE:3136755.3143004, Dhall:2018}. In 2017, the winner of the sub-challenge proposed fused deep models based on CNNs and trained on facial regions and entire images~\cite{Tan:2017:GER:3136755.3143008}. A deep hybrid network~\cite{Guo:2017:GER:3136755.3143017} using image scene, faces and skeletons attained the second place. 
  In 2018, the top performance of the sub-challenge was attained with a deep hybrid network~\cite{Guo:2018:GER:3242969.3264990} based on faces, scenes, skeletons, and visual attentions. Cascade attention networks~\cite{Wang:2018:CAN:3242969.3264991} based on face, body and image cues attained the second place and a four-stream deep network~\cite{Khan:2018:GER:3242969.3264987} consisting of the face-location aware global stream, the multi-scale face stream, a global blurred stream and a global stream attained the third place.

\subsection{Event Recognition}
With abundance of applications such as video surveillance and content-based video retrieval~\cite{Tzelepis:2016:EMP:3004988.3005088}, solutions to the problem of event recognition have evolved from using hand-engineered features to deep models for both videos~\cite{1570899, 5539870, Koller2012FEIFEI} and static images~\cite{Li07what,7301329, 7298768, 8296810}. Event recognition using static images is more challenging than using video because of the lack of motion information~\cite{Wang15}. The interest in event recognition from static images has increased due to the explosive growth of web images, driven primarily by online photo sharing services such as Flickr and Instagram. Event recognition is challenging because behaviors of interest can have a complex temporal structure. For example, a wedding event is characterized by behaviors that occur at different time, such as walking the bride, dancing, and flower throwing. Even though there are only 8 classes, each class encompasses many behaviors, which are visually very different from each other.  

In~\cite{Li07what}, an aggregate model that jointly infers the classes of event, scene and objects from low-level features of images was proposed.  Wang \etal ~\cite{Wang15} proposed a CNN which extracts useful information for event understanding from objects and scenes, and attained the first place in the task of cultural event recognition at the ChaLearn Looking at People (LAP) Challenge~\cite{7301329} in 2015. A framework that discovers event concept attributes from the web and use them to extract semantic features from images and classify them into social event categories was proposed in~\cite{ahsan2017complex}.

\begin{figure*}
\begin{center}
   \includegraphics[width=.81\linewidth]{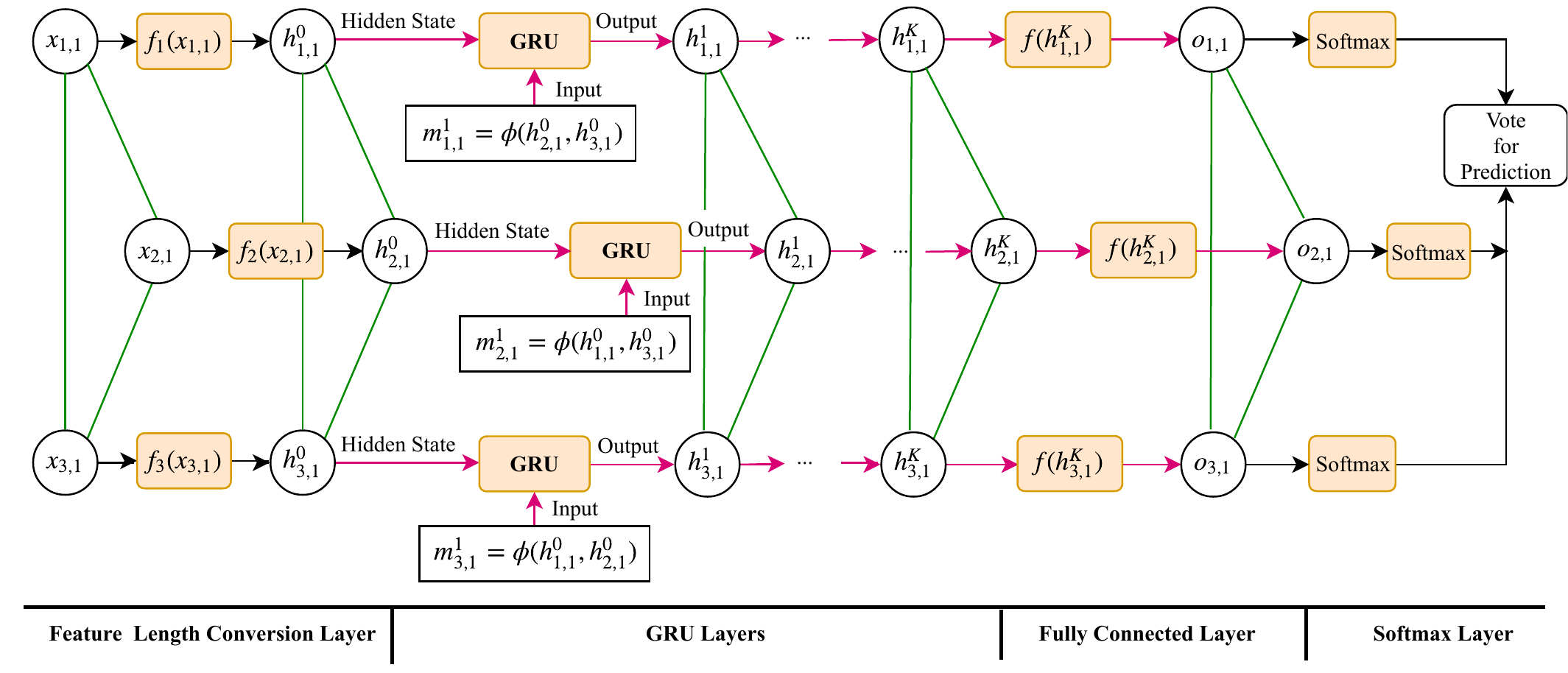}
\end{center}
   \caption{Illustration of a graph with 3 nodes. The feature vectors associated to the nodes are $x_{1,1}$, $x_{2,1}$ and $x_{3,1}$. The single layer neural networks used to convert the features from different cues to vectors of the same size  are $f_1$, $f_2$ and $f_3$. The resulting fixed-length vectors are $h_{1,1}^0$, $h_{2,1}^0$ and $h_{3,1}^0$. At each time step, GRUs take both the previous hidden states $h_{1,1}^{k-1}$, $h_{2,1}^{k-1}$, $h_{3,1}^{k-1}$ and the messages $m_{1,1}^k$, $m_{2,1}^k$, and $m_{3,1}^k$ as inputs, and output the updated hidden states $h_{1,1}^{k}$, $h_{2,1}^{k}$ and $h_{3,1}^{k}$. After $K$ time steps, the hidden states are fed to a fully connected layer to output class scores $o_{1,1}$, $o_{2,1}$ and $o_{3,1}$. The Softmax layer normalizes the class scores between 0 to 1, and majority voting over the nodes determines the final prediction.} 
\label{fig2}
\end{figure*}

\section{Proposed Graph Neural Network}
\label{GNN}
Motivated by previous works~\cite{7298768, Li07what, Wang15} that show image understanding benefits from the extraction of information from multiple cues, a GNN-based model is designed to jointly learn feature representations from multiple cues.

Given an image $I$, assume that there are $T$ different cue types of interest for a certain image understanding task. For example, Figure~\ref{fig1} illustrates $T=4$ cue types, namely, facial cues, human body cues, object cues and whole image cues. For each cue type $i$, $N_i$ features are extracted using deep models. For example, for the facial cues, $N_i$ may correspond to the number of detected faces in the image. The feature extraction operation for the $i$th cue is defined as 

\begin{equation} \label{eq2}
X_{i} = \psi_i(I),
\end{equation}
where, $X_{i} = [x_{i,1}, \ldots, x_{i,N_i}] \in{\mathbb{R}^{{L_i}\times{N_i}}}$ and $\psi_i$ denotes the set of ${L_i}$-dimensional features and the feature extractor operator corresponding to the $i$th cue type, respectively. For example, for facial cues, a candidate for $\psi_i$ may be an operator that detects face patches in the image and aligns them, runs the face patches through a fine-tuned VGG-FACE model~\cite{Parkhi15} and extracts the outputs from the fully-connected layer \textit{fc7} to generate features. 

To build the complete graph, each feature $x_{i,j}$ represents a node and every pair of distinct nodes is connected by an undirected edge. Note that $N_i$ may change across different images, for example, the number of faces changes across images, and therefore, every image has their own graph morphology. Since the feature length $L_i$ depends on the cue type, a function $f_i(\cdot)$ that converts the features to fixed-size vectors is needed. Although there are many options for the implementation of $f_i(\cdot)$, in this paper, the function is implemented with a single layer neural network as follows

\begin{equation} \label{eq3}
h_{i,j}^{0} = f_i(x_{i,j})=\text{ReLU}(W_i x_{i,j}+b_i),
\end{equation}
where $h_{i,j}^{0} \in \mathbb{R}^{{L_h}}$ is the fixed-length feature vector associated to $x_{i,j}$, $W_i\in{\mathbb{R}^{L_h\times{L_i}}}$ and $b_i\in \mathbb{R}^{{L_h}}$ are the cue-type-specific weight matrix and bias, respectively. The vectors $h_{i,j}^0$ will hereafter be referred to as the hidden states of the nodes. Note that $W_i$ and $b_i$ are shared across nodes corresponding to the same cue-type and ReLU can be replaced with other functions. 

The crucial idea of GNNs is that the vectors $h_{i,j}^{0}$ are iteratively updated by trainable nonlinear functions that depend on the hidden states of the neighbor nodes. This is accomplished by a GRU model in this paper. At every time step $k$, the hidden states are updated with a new $h_{i,j}^{k}$. Since the fixed-size features are the initial state input to the GRU, $L_h$ is also the number of hidden units in the GRU. As shown in Figure~\ref{fig2}, a GRU unit takes the previous hidden state of the node $h_{i,j}^{k-1}$ and a message $m_{i,j}^{k}$ as input at each iteration, and outputs a new hidden state $h_{i,j}^{k}$. The message $m_{i,j}^k$, generated at time step $k$, is the aggregation of messages from the neighbors of the node, and is defined by the aggregation function $\phi(\cdot)$ as

\begin{eqnarray}\label{eq4}
m_{i,j}^{k}&=&\phi(\{h_{q,p}^{k-1}\mid \forall (q, p), (q,p)\neq (i,j)\}),\\
&=&\sum_{\substack{q,p\\
                  (q,p)\neq (i,j)}}
        W^e_qh_{q,p}^{k-1},
\end{eqnarray}
where $W_q^e\in\mathbb{R}^{L_h\times{L_h}}$ is the weight matrix associated to the neighbors whose cue type is $q$. Note that the neighbors are all the other nodes since the graph is complete. The cue-dependent matrices $W_q^e$ are learned during training.

The computations within the GRU, which allow the network to adaptively reset or update its memory content, are formally expressed as follows:
\begin{equation}
\label{eq7}
\begin{aligned}
z_{i,j}^k&=\sigma(W_z m_{i,j}^k+U_zh_{i,j}^{k-1}),\\
r_{i,j}^k&=\sigma(W_rm_{i,j}^k+U_rh_{i,j}^{k-1}),\\
\tilde{h}_{i,j}^k&=\text{tanh}(W_hm_{i,j}^k+U_h(r_{i,j}^k\odot{h_{i,j}^{k-1}})),\\
h_{i,j}^k&=(1-z_{i,j}^k)\odot{h_{i,j}^{k-1}}+z_{i,j}^k\odot{\tilde{h}_{i,j}^k},
\end{aligned}
\end{equation}
where $r_{i,j}^k$ and $z_{i,j}^k$ are the reset and update gates, $\tilde{h}_{i,j}^k$ is the candidate memory content, $\sigma(\cdot)$ is the logistic sigmoid function, and $\odot$ denotes the element-wise multiplication operation, and matrices $W_z$, $W_r$, $W_h$, $U_z$, $U_r$, and $U_h$ are model parameters. The update gate $z_{i,j}^k$ controls how much of the previous memory content is to be forgotten and how much of the candidate memory content is to be added. The model parameters of the GRU are shared across all nodes, thus providing an explicit control on the number of parameters. After training the GRU for $K$ time steps, all the nodes have learned from their neighbors during $K$ iterations. Note that the functions that define the update of the hidden states specify
a propagation model of information inside the graph.

The final stage of the GNN consists in pushing the last hidden states through a fully-connected (FC) layer followed by a Softmax layer to generate the class probabilities. The total number of classes is denoted as $C$. The FC layer is represented with the function $f(\cdot)$, which is defined as
\begin{equation} \label{eq8}
o_{i,j} = f(h_{i,j}^K)=Wh_{i,j}^K+b,
\end{equation}
where $W\in{R^{C\times{L_h}}}$ and $b\in{R^{{C}}}$ are the weights and bias term of the FC layer and are the same for all the nodes in the network. The class probabilities are generated by the Softmax layer as follows,

\begin{equation} \label{eq9}
p_{i,j}^c = \frac{e^{W_{(c)}h_{i,j}^K+b_{(c)}}}{ \sum_{l=1}^{C} e^{W_{(l)}h_{i,j}^K+b_{(l)}}},
\end{equation}
where $p_{i,j}^c$ is the probability for class $c$, $W_{(c)}$ is the $c$th row of $W$ and $b_{(c)}$ is the $c$th component of $b$. The predicted class of a node is the class with the largest probability, and the final prediction of the GNN is computed by using majority voting over the class predictions of the nodes. Figure~\ref{fig2} illustrates the structure of the proposed GNN.



The GNN is trained using backpropagation through time and the cross entropy loss function for multiple cues, which is defined, for each training sample, as
\begin{equation} \label{eq10}
L = -\frac{1}{\sum_{i} N_i}\sum_{i,j}\sum_{c}{y_c log(p_{i,j}^c)},
\end{equation}
where $y_c$ is the ground-truth for class $c$.


\section{Experiments}
\label{experiments}
In this section, the GroupEmoW database is introduced. Details of the implementation of the proposed GNN and comparisons with baseline and state-of-the-art methods are also provided. 

\subsection{Datasets}

\subsubsection{GroupEmoW: A New GER Dataset}
Datasets are crucial for building deep learning models. Even though there are many images of groups of people on social media and a strong interest in GER, labeled data is still scarce. In this paper, a new group-level emotion dataset in the wild, referred to as GroupEmoW, is introduced. The images are collected from Google, Baidu, Bing, and Flickr by searching for keywords related to social events, such as funeral, birthday, protest, conference, meeting, wedding, etc. Collected images form an in-the-wild dataset, with different image resolutions. The labeling task was performed by trained human annotators, including professors and students. Each image is labelled by 5 annotators, and the ground-truth is determined by consensus. Images are removed from the dataset if a consensus is not reached. 
\begin{table}
\begin{center}
\begin{tabular}{l|c|c|c|c}
\hline
Dataset & Partition & Neg & Neu & Pos  \\
\hline\hline
\multirow{3}{*}{\shortstack[l]{Group Affect\\Database 2.0~\cite{Dhall:2018}}}&Train& 2759 &3080&3977\\
                                                                                 & Val&1231&1368&1747\\
&Test&1266&916&829\\
\hline
MultiEmoVA~\cite{7284862}&--&68& 73& 109\\
\hline

\multirow{3}{*}{GroupEmoW}&Train&3019&3463&4645\\
           &Val&861&990&1327\\
           &Test&431&494&664\\
\hline
\end{tabular}
\end{center}
\caption{Dataset distribution of the proposed GroupEmoW dataset and currently available datasets for GER, where column names Neg, Neu and Pos correspond to negative, neutral and positive, respectively.}\label{table1}
\end{table}

\begin{figure}[t]
\begin{center}
   \includegraphics[width=0.8\linewidth]{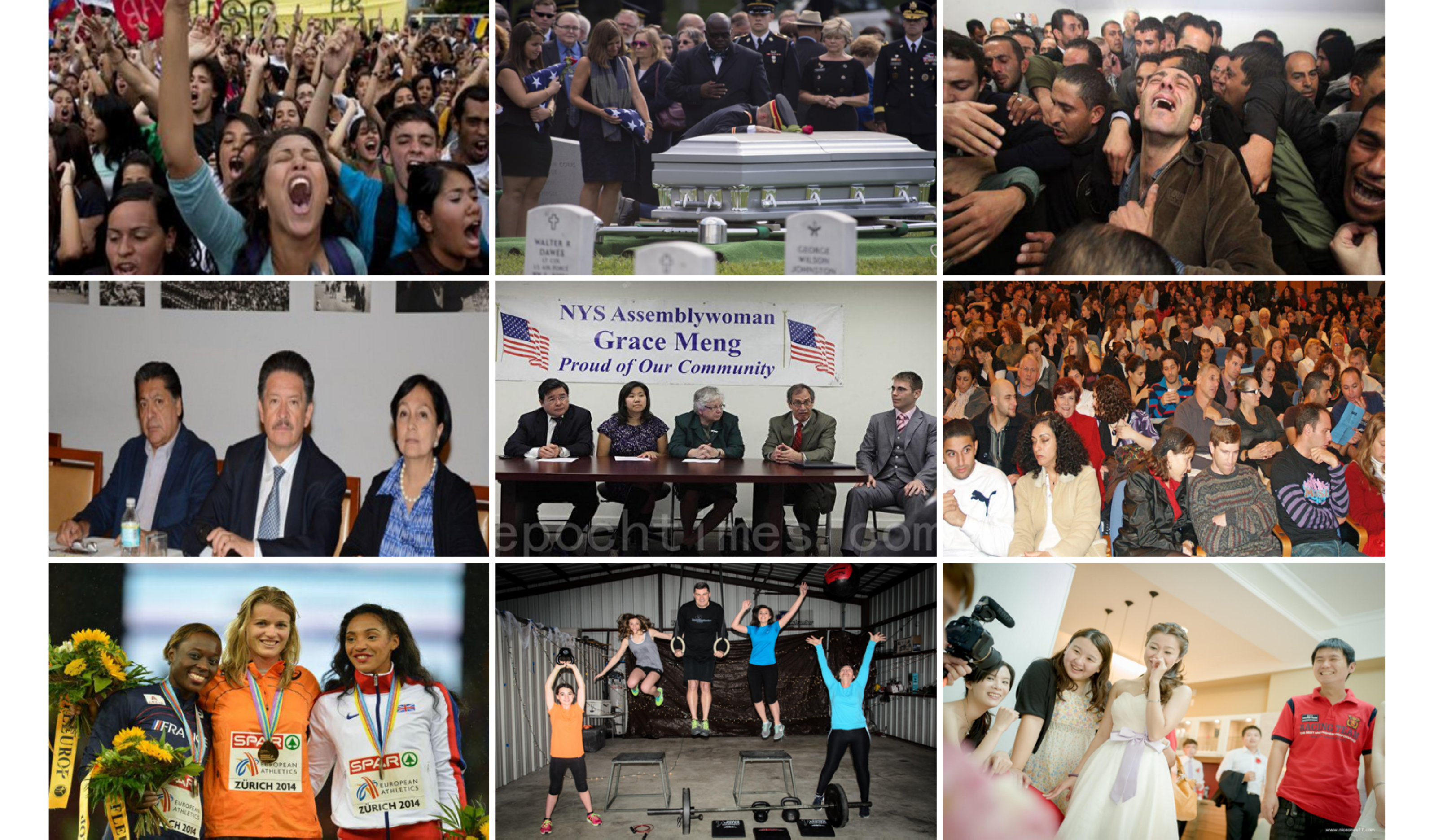}
\end{center}
   \caption{GroupEmoW samples. First row: negative valence state. Middle row: neural valence state. Last row: positive valence state.}
\label{fig3}
\end{figure}

The collective emotion of the images are labeled between negative, neutral, and positive valence states. The total number of $15,894$ images in the GroupEmoW database is divided into train, validation and test sets with $11,127$, $3,178$ and $1,589$ images, respectively. The distribution of samples and comparison with currently available datasets for the GER problem are shown in Table~\ref{table1}. Sample images of the GroupEmoW database are shown in Figure~\ref{fig3}.


\subsubsection{Group Affect Database 2.0}
 The Group Affect Database $2.0$~\cite{Dhal15b} contains $9,816$, $4,346$ and $3,011$ images in the train, validation and test sets, respectively. These images are associated to social events, such as convocations, marriages, parties, meetings, funerals, protests, etc. This is the dataset employed by the GER sub-challenge of the Emotion Recognition in the Wild (EmotiW) Grand Challenge~\cite{Dhall:2018}. The labels of train and validation sets are provided while the labels of the test set are unknown. The size of the Group Affect Database $2.0$ was increased from $6,467$ in $2017$ to $17,173$ in $2018$. 
 
 \subsubsection{Social Event Image Dataset (SocEID)}
 The Social Event Image Dataset (SocEID)~\cite{ahsan2017complex} is a large-scale dataset that consists of $37,000$ images belonging to $8$ event classes (birthdays, graduations, weddings, marathons/races, protests, parades, soccer matches and concerts). It was collected by querying Instagram and Flickr with tags related to the event of interest. This dataset also contains some relevant images from the NUS-WIDE dataset~\cite{Chua:2009:NRW:1646396.1646452} and the Social Event Classification subtask from MediaEval $2013$~\cite{quteprints66957}. SocEID contains $27,718$ and $9,254$ samples in the train and test sets, respectively. 
 
\subsection{Implementation and Results}
Three baseline methods are proposed as follows:

(1) A fine-tuned CNN model based on whole images, referred to as CNN-Image. The selected pre-trained CNN is SE-ResNet-50~\cite{DBLP:journals/corr/abs-1709-01507}, which is a $50$-layer version of the SENet-154 model~\cite{DBLP:journals/corr/abs-1709-01507}, which was trained on the ImageNet-1K database and achieved the highest accuracy in the ILSVRC 2017 image classification challenge\footnote{The SE-ResNet-50 and SENet-154 pre-trained models are downloaded from https://github.com/hujie-frank/SENet.}. All the learning parameters are adopted from the original model, with the exception of the size of the last FC layer, which is set the same as the number of classes of the problem of interest (3 for GER and 8 for event recognition), and the learning rate, which is initialized to $0.0005$.

(2) GRU and long short-term memory (LSTM) models trained on single cue types. For example, for facial cues, these models treat facial features within one image as one input sequence. The output of the RNN, either GRU or LSTM, is connected to an FC layer followed by a Softmax layer to generate predictions. The learning rate and length of the hidden state vectors of these models are set to 0.0001 and 128, respectively. The GRUs trained for faces and objects are referred to as GRU-Face and GRU-Object, respectively. The LSTM models trained for faces and objects are referred to as LSTM-Face and LSTM-Object, respectively. Even though GRU and LSTM models can handle variable-length input sequences, each input must have the same feature size, which lead us to train GRU and LSTM models only on single cues.

(3) Additional baselines referred to as CNN-VGG-F and CNN-Skeleton are described in Sections \ref{Sec-GroupEmoW} and \ref{Sec-GroupAffect}, respectively.

In addition, the performance of the proposed GNN is also compared with state-of-the-art methods for GER and event recognition. All experiments are performed 10 times, using 20 epochs each time. For the GroupEmoW database, the model at the epoch with the highest average accuracy on the validation set is selected. Results on the test set using the selected model are reported. For SocEID, given that the dataset is divided into two partitions only, and for the Group Affect Database 2.0, given that the test labels are unknown, the model at the epoch with the highest average accuracy on the training set is selected. Results on the validation set are reported for the Group Affect Database 2.0., while results on the test set are reported for SocEID. For all the GNN models, the learning rate, the number of time steps $K$, and the length of the hidden state vectors $L_h$ are set to 0.0001, 4, and 128, respectively. The learning rate was selected using grid search in \{0.00001, 0.0001, 0.001, 0.01\}. The performance metric used for evaluation is the classification accuracy.

\subsubsection{Experiments on the GroupEmoW Database}
\label{Sec-GroupEmoW}
\begin{figure}[t]
\begin{center}
   \includegraphics[width=.87\linewidth]{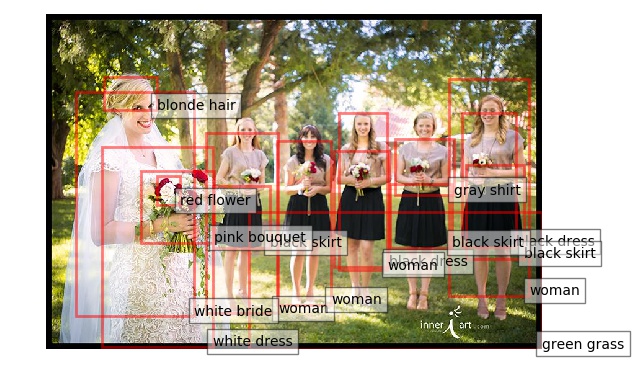}
\end{center}
   \caption{Salient object areas within one image.}
\label{fig4}
\end{figure}
Three cue types, namely, facial, object and whole image cues are explored for the GER task using the GroupEmoW database. Face patches are extracted and aligned using MTCNN~\cite{7553523}. A VGG-FACE model~\cite{Parkhi15} initially trained on 2.7M images for face recognition is fine-tuned in the same way as described in~\cite{Guo:2018:GER:3242969.3264990} but using the training set of the GroupEmoW database.  Once fine-tuned, the features of the FC layer \textit{fc7} are extracted from each face patch and used as input to the GRU, LSTM, and GNN models. A baseline method, referred to as CNN VGG-F and described in ~\cite{Guo:2018:GER:3242969.3264990}, is implemented by running the face patches of an image through the fine-tuned VGG-FACE model and averaging the generated class probabilities across faces to finally select the class with the largest average class probability.

Let $i=1$ be the index assigned to the facial cue type. The number of face feature vectors $N_1$ extracted from an image is restricted to be less or equal than $N_1^{\text{max}}$. During training, to ensure that all of the faces from most of the training images are selected, $N_1^{\text{max}}$ is set to 16 since $84.68\%$ of the training images in the GroupEmoW database contain less than $16$ faces. If more than $N_1^{\text{max}}=16$ faces are detected in an image, then $N_1^{\text{max}}=16$ faces are randomly selected to extract features from them. During testing, $N_1^{\text{max}}$ is set to 48 since $98.46\%$ of the testing images contain less than $48$ faces. If more than $48$ faces are detected in a testing image, the first $48$ faces to be detected are selected. Therefore, faces are selected in a deterministic fashion during testing. The reason for using a smaller $N_1^{\text{max}}$ for training than for testing is to prevent images with a large number of faces to excessively influence the learning of the network.

Facial features within one image are treated as one input sequence by the GRU and LSTM models. Therefore, the maximum sequence length is $16$ for training and $48$ for testing. The number of time steps for the GRU and LSTM models are equal to the input sequence length, which is the number of extracted face patches $N_1$. 

For the object cues, the attention mechanism  proposed in~\cite{DBLP:journals/corr/AndersonHBTJGZ17} is used to extract the salient objects. The SENet-154 model~\cite{DBLP:journals/corr/abs-1709-01507} trained on the ImageNet-1K database is employed to extract a 2048-dimensional feature representation for each salient object by using the output of layer \textit{pool5/7x7\_s1}. As shown in Figure~\ref{fig4}, the attention mechanism is able to detect salient objects, such as humans, bouquet and grass. The areas detected by the attention mechanism are sorted by the confidence of the predictions. Let $i=2$ be the index of the object cues. The number of feature vectors for the salient objects is restricted to be less or equal than $N_2^{\text{max}}$; therefore, if more than $N_2^{\text{max}}$ salient objects are detected by the attention mechanism, then only the salient objects with the top $N_2^{\text{max}}$ scores are selected for feature extraction. The value of $N_2^{\text{max}}$ is set to 16 for the experiments in this section. 

For  the  whole image  cues, in order to show that the proposed GNN is able to handle features of different length, an Inception-V2~\cite{DBLP:journals/corr/SzegedyVISW15} model pre-trained on the ImageNet-1K database is fine-tuned as described in~\cite{Guo:2018:GER:3242969.3264990} but using the training set of the GroupEmoW database. Once fine-tuned, the features of the \textit{global\_pool} layer with dimension $1024$ are extracted and used as input to the GNN models.

The performance of the GNN model is evaluated by progressively adding cues of different type. First, the performance of the GNN using facial cues only, referred to as GNN-Face, and object cues only, referred to as GNN-Object, is evaluated. Next, the performance of the GNN that uses both object and facial cues, referred to as GNN F+O, is evaluated. The last model to be evaluated is the GNN that uses face, object and whole image cues, referred to as GNN F+O+I. Results shown in Table~\ref{tab_GroupEmoW} demonstrate that the proposed GNN F+O+I model outperforms the baseline methods. Each cue type adds information that is needed to improve the overall accuracy. Note that both GRU-Face and LSTM-Face slightly outperform GNN-Face, while GNN-Object outperforms both GRU-Object and LSTM-Object. This may be due to the fact that similarity between face patches is much higher than similarity between salient objects, and therefore, the task of predicting group-level emotion from faces may benefit from a simpler model. Instead, relations between salient objects are more semantic and may need more elaborate models.   

\begin{table}
\begin{center}
\begin{tabular}{l|c|c|c|c|c}
\hline
Method & Avg\_V&Max & Min & Med & Avg  \\
\hline\hline
CNN-Image&80.14&82.38&79.11&81.25&81.22\\
CNN-VGG-F  &83.17&82.52&81.95&82.27&82.26\\
GRU-Face &85.66&85.65&84.83&85.15&85.28\\
LSTM-Face &85.58&85.27&84.45&84.70&84.86\\
\textbf{GNN-Face} &85.54&85.02&84.14&84.64&84.68\\
\hline
GRU-Object &85.38&85.58&83.95&84.58&84.83\\
LSTM-Object &85.25&85.52&83.95&84.77&84.92\\
\textbf{GNN-Object} &85.93&86.21&85.08&85.71&85.66\\
\hline
\textbf{GNN F+O} &89.71&89.80&88.35&89.03&89.06\\
\textbf{GNN F+O+I}&\textbf{89.79}&\textbf{89.93}&\textbf{88.60}&\textbf{89.11}&\textbf{89.14}\\
\hline
\end{tabular}
\end{center}

\caption{Experimental results on the GroupEmoW dataset. Avg\_V refers to the average accuracy on the validation set, while Max, Min, Med, Avg are maximal, minimal, median and average accuracy on the test set. F, O, and I refer to face, object and whole image cues, respectively.}\label{tab_GroupEmoW}
\end{table}


\subsubsection{Experiments on the Group Affect Database 2.0}
\label{Sec-GroupAffect}
\begin{figure}[t]
\begin{center}
   \includegraphics[width=.95\linewidth]{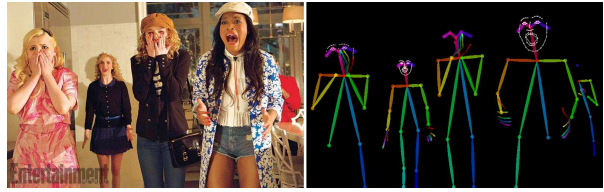}
\end{center}
   \caption{A sample image for the negative valence state from the Group Affect Database 2.0. and its corresponding skeleton image.}
\label{fig5}
\end{figure}

In addition to the three cues used for the GroupEmoW Database, skeleton cues are also used for the Group Affect Database 2.0. Skeleton images have been used in~\cite{Guo:2017:GER:3136755.3143017, Guo:2018:GER:3242969.3264990} for group level emotion recognition and offer crucial information related to people layout and postures. Skeleton images only contain the landmarks of the faces and limbs and their connections (Figure~\ref{fig5}). OpenPose~\cite{cao2017realtime,simon2017hand,wei2016cpm} is used to extract skeleton images in the same way as described in ~\cite{Guo:2017:GER:3136755.3143017, Guo:2018:GER:3242969.3264990}. The SE-ResNet-50 is fine-tuned on skeleton images in the same way as described in~\cite{Guo:2018:GER:3242969.3264990}. Once fine-tuned, the features of the \textit{pool5/7x7\_s1} layer are extracted from each skeleton image and used as one of the inputs of the GNN model. The CNN model trained on skeleton images, described in~\cite{Guo:2018:GER:3242969.3264990}, and referred to as CNN-Skeleton, is used as a baseline method in Table~\ref{tab3}.

As in Section \ref{Sec-GroupEmoW}, the number of features for the facial and object cues is also restricted to be less or equal than $N_1^{\text{max}}$ and $N_2^{\text{max}}$, respectively. During training, to ensure that all of the faces from most of the training images are selected, $N_1^{\text{max}}$ is set to $16$ since $86.72\%$ of the training images in the Group Affect Database 2.0 contain less than $16$ faces. During testing, $N_1^{\text{max}}$ is set to 48 since $98.58\%$ of the testing images contain less than $48$ faces. For the object cues, $N_2^{\text{max}}$ is set to 16 for both training and testing.


As in Section \ref{Sec-GroupEmoW}, the performance of the GNN is evaluated by progressively adding cues of different type. Other than the comparisons with the baseline models in Table~\ref{tab3}, GNN is also compared to state-of-the-art methods. Since the methods described in~\cite{Dhall:2018, Khan:2018:GER:3242969.3264987, Wang:2018:CAN:3242969.3264991, Guo:2018:GER:3242969.3264990} report their best predictions across different experiments on the validation set, their results are placed in the column that reports the maximum accuracy in  Table~\ref{tab3}. We are unable to evaluate the performance of the proposed GNN on the test set since the test labels are unavailable. In terms of average and median accuracy, experimental results show that GNN-based models outperform GRU and LSTM models trained on single cues. The proposed model that exploits face, object, whole image and skeleton cues, referred to as GNN F+O+I+S, outperforms all the state-of-the-art methods in Table~\ref{tab3}, except the model in~\cite{Wang:2018:CAN:3242969.3264991}, which attains high accuracy on the validation dataset but lower accuracy than the model in~\cite{Guo:2018:GER:3242969.3264990} on the test set.

\begin{table}
\begin{center}
\begin{tabular}{l|c|c|c|c}
\hline
Method & Max & Min & Median & Avg  \\
\hline\hline
CNN-Image~\cite{Guo:2018:GER:3242969.3264990} &68.16&--&--&--\\
CNN-Skeleton~\cite{Guo:2018:GER:3242969.3264990} &64.42&--&--&--\\
CNN-VGG-F~\cite{Guo:2018:GER:3242969.3264990} & 68.28&-- &-- &--\\
GRU-Face &75.34&74.68&74.99&75.05\\
LSTM-Face &75.45&74.22&75.06&75.03\\
\textbf{GNN-Face} &75.48&73.96&75.24&75.00\\
\hline
GRU-Object &68.45&65.09&66.78&66.89\\
LSTM-Object &67.39&66.06&66.68&66.77\\
\textbf{GNN-Object} &69.16&67.76&68.34&68.32\\
\hline
Inception-Img~\cite{Dhall:2018}&65.00&--&--&--\\
Multi-Models~\cite{Khan:2018:GER:3242969.3264987}&78.39&--&--&--\\
Multi-Models~\cite{Wang:2018:CAN:3242969.3264991}&86.90&--&--&--\\
Multi-Models~\cite{Guo:2018:GER:3242969.3264990}&78.98&--&--&--\\
\textbf{GNN F+O} &78.34&76.32&77.58&77.83\\
\textbf{GNN F+O+I}&78.87&76.65&77.97&77.96\\
\textbf{GNN F+O+I+S}&\textbf{79.08}&\textbf{77.09}&\textbf{78.00}&\textbf{78.16}\\
\hline
\end{tabular}
\end{center}
\caption{Comparison with baseline and state-of-the-art methods using the validation set of the Group Affect Database 2.0 dataset. Note that the multi-model method in ~\cite{Wang:2018:CAN:3242969.3264991} attains high metrics in the validation set but the performance on the test set is lower than that of the method in~\cite{Guo:2018:GER:3242969.3264990}. F, O, I, and S refer to face, object, whole image, and skeleton cues respectively.}\label{tab3}
\end{table}

\subsubsection{Experiments on the SocEID Database}
The same cues used in Section~\ref{Sec-GroupEmoW} are employed for the event recognition task, with the exception of the facial cues, which are replaced by human body cues since faces are not as important as human bodies when it comes to recognizing activities and scene categories. Human body bounding boxes are detected and cropped in the following way: face and body keypoints are first detected using OpenPose~\cite{cao2017realtime,simon2017hand,wei2016cpm}, the width and height of the bounding boxes for the detected keypoints are calculated, and then increased by $20\%$. Any bounding box region that lies outside the image is cropped to fit within the image.

Since the average number of humans in the SocEid dataset is only $2$, human body cues-based CNNs are not trained in this paper. Instead, the human body features are extracted using the output of the \textit{pool5/7x7\_s1} layer from the pre-trained SENet-154 model.  The number of human bodies to be extracted from a single image is restricted to be less or equal than 16 for both training and testing.

Features for the whole image and salient object cues are extracted in the same way as described in Section~\ref{Sec-GroupEmoW}. The number of salient objects is restricted to be less or equal than $16$. As in Section \ref{Sec-GroupEmoW}, the performance of the GNN is evaluated by progressively adding cues of different type. In  Table~\ref{tab4}, GNN O+H refers to the model that exploits object and human body cues, while GNN O+H+I refers to the model that uses object, human body, and whole image cues. Table~\ref{tab4} shows that the proposed models outperform baseline and state-of-the-art methods.

\begin{table}
\begin{center}
\begin{tabular}{l|c|c|c|c}
\hline
Method & Max & Min & Median & Avg  \\
\hline\hline
CNN-Image &89.18&87.86&88.66&88.62\\
GRU-Object &90.12&90.27&90.67&90.69\\
LSTM-Object &90.90&90.36&90.71&90.67\\
\textbf{GNN-Object} &91.47&90.79&91.27&91.17\\
\hline
AlexNet-fc7~\cite{ahsan2017complex}&--&--&--&86.42\\
Event concept~\cite{ahsan2017complex}&--&--&--&85.39\\
\textbf{GNN O+H} &91.96&90.73&91.38&91.33\\
\textbf{GNN O+H+I}&\textbf{92.09}&\textbf{91.27}&\textbf{91.52}&\textbf{91.61}\\

\hline
\end{tabular}
\end{center}
\caption{Experimental results on the SocEID dataset. O, H, and I refer to object, human and whole image cues, respectively.}\label{tab4}
\end{table}



\section{Discussion and Future work}
\label{discussion}
The success of CNNs is partially owed to their ability to exploit local information, by enforcing a local connectivity pattern between neurons, and to aggregate and synthesize those local attributes in the upper layers of the network to learn high-level representations. However, there is a need to transition from models that are able to extract and aggregate local attributes for tasks such as object recognition and segmentation to models that are able to extract and aggregate local attributes for reaching a complete understanding of images. Progress in that direction has been attained with attention mechanisms that help models focus on the salient areas of the image. Traditional feature fusion approaches used to aggregate features from those salient areas ignore the relations between features and their ability to learn from each other. Similarly, RNN-based approaches ignore the spatial relations between salient areas, which are better described as a set than as a sequence. \textbf{The application of GNNs to image understanding tasks effectively learns feature representations for the salient regions by exchanging information between the graph nodes. The design of the GNN allows substantial weight sharing, which helps to avoid overfitting}. 

There is no guarantee that all the extracted regions from the image provide relevant information for the task of interest, some of the regions may be uncorrelated or may introduce noise. Therefore, building a complete graph may not be optimal. \textbf{Future work will address the problem of efficiently connecting the graph nodes. In the future, we may also consider jointly learning the parameters of the GNN and the CNNs used for feature extraction in an end-to-end fashion}. The proposed method can be applied to other image understanding tasks that involve aggregating information from multiple cues, such as image captioning, visual grounding, and visual question answering.

\section{Conclusion}
\label{conclusion}
A GNN-based framework for image understanding from multiple cues and a new database for the GER problem were presented in this paper. Image understanding not only refers to identifying objects in an image but also to learning the underlying interactions and relations between those objects. Exploiting those relations during the feature learning and prediction stages is achieved with GNNs by propagating node messages through the graph and aggregating the results. A variety of experimental results show that the proposed model achieves state-of-the-art performance on GER and event recognition tasks.

\section{Acknowledgements}
The work is supported by the National Science Foundation under Grant No.~{$1319598$}.

{\small
\bibliographystyle{ieee}
\bibliography{egbib}
}

\end{document}